%% file: main.tex
\documentclass[11pt,a4paper]{article}
\usepackage{times,latexsym}
\usepackage{url}
\usepackage[T1]{fontenc}

    \usepackage[acceptedWithA]{tacl2018v2}
\usepackage{rotating}
\usepackage{blindtext}
\usepackage{amsmath}
\usepackage{amsfonts}
\usepackage{graphicx}
\usepackage{mathtools}
\usepackage{booktabs}
\usepackage{relsize}
\usepackage{float}
\usepackage{graphics}
\usepackage{multirow}
\usepackage{lipsum}
\usepackage{placeins}

\usepackage[utf8]{inputenc}
\DeclareUnicodeCharacter{01B0}{\`u}
\DeclareUnicodeCharacter{01A1}{\`o}

\usepackage{threeparttable}
\usepackage{booktabs, makecell}

\usepackage[title]{appendix}
\usepackage[cal=cm]{mathalfa}
\usepackage[colorinlistoftodos]{todonotes}
\usepackage{makecell,multirow}
\usepackage[style=base]{caption}
\usepackage[export]{adjustbox}
\usepackage[ruled,vlined]{algorithm2e}
\usepackage[subtle]{savetrees}
\newcommand{\ignore}[1]{}

\usepackage{xspace,mfirstuc,tabulary}

\newif\iftaclinstructions
\taclinstructionsfalse 
\iftaclinstructions
\renewcommand{\confidential}{}
\renewcommand{\anonsubtext}{(No author info supplied here, for consistency with
TACL-submission anonymization requirements)}
\newcommand{\instr}
\fi

\iftaclpubformat 

\else

\fi

\title{Recursive Non-Autoregressive Graph-to-Graph Transformer \\
  for Dependency Parsing with Iterative Refinement}

\author{Alireza Mohammadshahi \\ 
  Idiap Research Institute / EPFL \\
  \texttt{alireza.mohammadshahi@idiap.ch}
  \And 
  James Henderson\\
  Idiap Research Institute \\
  \texttt{james.henderson@idiap.ch}
  \\ }

\date{}

\begin{document}
\maketitle
\interfootnotelinepenalty=10000

\input{abstract.tex}

\input{intro.tex}
\input{model.tex}

\input{implement.tex}

\input{results.tex}
\input{conclusion.tex}

\input{ack.tex}

\newpage

\bibliographystyle{acl_natbib}
\bibliography{anthology,acl2020}

\newpage
\renewcommand\thesection{\Alph{section}}
\renewcommand\thesubsection{\thesection.\Alph{subsection}}
\input{supplementary.tex}

\end{document}

%% file: abstract.tex
\begin{abstract}
  We propose the Recursive Non-autoregressive Graph-to-Graph Transformer architecture (RNGTr) for the iterative refinement of arbitrary graphs through the recursive application of a non-autoregressive Graph-to-Graph Transformer and apply it to syntactic dependency parsing.
  We demonstrate the power and effectiveness of RNGTr on several dependency corpora, using a refinement model pre-trained with BERT.
  We also introduce Syntactic Transformer (SynTr), a non-recursive parser similar to our refinement model.  RNGTr can improve the accuracy of a variety of initial parsers on 13 languages from the Universal Dependencies Treebanks, English and Chinese Penn Treebanks, and the German CoNLL2009 corpus, even improving over the new state-of-the-art results achieved by SynTr, significantly improving the state-of-the-art for all corpora tested.
\end{abstract}

%% file: intro.tex
\section{Introduction}

Self-attention models, such as Transformer~\cite{vaswani2017attention}, have been hugely successful in a wide range of natural language processing (NLP) tasks, especially when combined with language-model pre-training, such as BERT~\cite{devlin2018bert}.  These architectures contain a stack of self-attention layers which can capture long-range dependencies over the input sequence, while still representing its sequential order using absolute position encodings. Alternatively, \newcite{shaw-etal-2018-self} proposes to define sequential order with relative position encodings, which are input to the self-attention functions.  

Recently \newcite{mohammadshahi2019graphtograph} extended this sequence input method to the input of arbitrary graph relations via the self-attention mechanism, and combined it with an attention-like function for graph relation prediction, resulting in their proposed Graph-to-Graph Transformer architecture (G2GTr).  They demonstrated the effectiveness of G2GTr for transition-based dependency parsing and its compatibility with pre-trained BERT~\cite{devlin2018bert}. 
This parsing model predicts one edge of the parse graph at a time, conditioning on the graph of previous edges, so it is an autoregressive model.

The G2GTr architecture could be used to predict all the edges of a graph in parallel, but such predictions are non-autoregressive. They thus cannot fully model the interactions between edges.  For sequence prediction, this problem has been addressed with non-autoregressive iterative refinement~\cite{novak2016iterative,lee-etal-2018-deterministic,awasthi2019parallel,lichtarge2018weakly}.  Interactions between different positions in the string are modelled by conditioning on a previous version of the same string.

In this paper, we propose a new graph prediction architecture which takes advantage of the full graph-to-graph functionality of G2GTr to apply a G2GTr model to refine the output graph recursively.  
This architecture predicts all edges of the graph in parallel, and is therefore non-autoregressive, but can still capture any between-edge dependency by conditioning on the previous version of the graph, like an auto-regressive model.  

This proposed Recursive Non-autoregressive Graph-to-Graph Transformer (RNGTr) architecture has three components.  First, an initialisation model computes an initial graph, which can be any given model for the task, even a trivial one.
Second, a G2GTr model takes the previous graph as input and predicts each edge of the target graph.  Third, a decoding algorithm finds the best graph given these edge predictions.  The second and third components are applied recursively to do iterative refinement of the output graph until some stopping criterion is met.
The final output graph is the graph output by the final decoding step.

The RNG Transformer architecture can be applied to any task with a sequence or graph as input and a graph over the same set of nodes as output.
We evaluate RNGTr on syntactic dependency parsing because it is a difficult structured prediction task, state-of-the-art initial parsers are extremely competitive, and there is little previous evidence that non-autoregressive models (as in graph-based dependency parsers) are not sufficient for this task.  We aim to show that capturing correlations between dependencies with non-autoregressive iterative refinement results in improvements, even in the challenging case of state-of-the-art dependency parsers.

The evaluation demonstrates improvements with several initial parsers, including previous state-of-the-art dependency parsers, and the empty parse.  We also introduce a strong Transformer-based dependency parser pre-trained with BERT~\cite{devlin2018bert}, called Syntactic Transformer (SynTr), using it both for our initial parser and as the basis of our refinement model.  Results on 13 languages from the Universal Dependencies Treebanks~\cite{11234/1-2895}, English and Chinese Penn Treebanks~\cite{marcus-etal-1993-building,xue-etal-2002-building}, and German CoNLL 2009 corpus~\cite{hajic-etal-2009-conll} show significant improvements over all initial parsers and the state-of-the-art.\footnote{Our implementation is available at: \url{https://github.com/idiap/g2g-transformer}}

In this paper, we make the following contributions:

\vspace{-1ex}
\begin{itemize}\addtolength{\itemsep}{-2ex}
\item We propose a novel architecture for the iterative refinement of arbitrary graphs
  (RNGTr) which combines non-autoregressive edge prediction with conditioning on the complete graph.
\item We propose a RNGTr model of syntactic dependency parsing.
\item We demonstrate significant improvements over the previous state-of-the-art dependency parsing results on Universal Dependency Treebanks, Penn Treebanks, and the German CoNLL 2009 corpus.
\end{itemize}

\section{Dependency Parsing} 

Syntactic dependency parsing is a critical component in a variety of natural language understanding tasks, such as semantic role labelling~\cite{marcheggiani-titov-2017-encoding}, machine translation~\cite{Chen_2017}, relation extraction~\cite{zhang-etal-2018-graph}, and natural language interfaces~\cite{pang2019improving}.  There are several approaches to compute the dependency tree.
\textit{Transition-based} parsers predict the dependency graph one edge at a time through a sequence of parsing actions \cite{yamada-matsumoto-2003-statistical,nivre-scholz-2004-deterministic,titov-henderson-2007-latent,zhang-nivre-2011-transition}.
As in our approach, \textit{transformation-based} \cite{satta-brill-1996-efficient} and \textit{corrective modeling} parsers use various methods (e.g.\ \cite{Knight05,hall-novak-2005-corrective,attardi-ciaramita-2007-tree,hennig2017dependency,zheng-2017-incremental}) to correct an initial parse.
We take a graph-based approach to this correction.
\textit{Graph-based} parsers~\cite{eisner-1996-three,mcdonald-etal-2005-online,koo2010efficient} compute scores for every possible dependency edge and then apply a decoding algorithm to find the highest scoring total tree.  Typically neural graph-based models consist of two components: an \textit{encoder} which learns context-dependent vector representations for the nodes of the dependency graph, and a \textit{decoder} that computes the dependency scores for each pair of nodes and then applies a decoding algorithm to find the highest-scoring dependency tree.

There are several approaches to capture correlations between dependency edges in graph-based models. In first-order models, such as Maximum Spanning Tree (MST)~\cite{edmonds1967optimum,Chu1965OnTS,mcdonald-etal-2005-non}, the score for an edge must be computed without being sure what other edges the model will choose. The model itself only imposes the discrete tree constraint between edges. In higher-order models~\cite{mcdonald-pereira-2006-online,carreras-2007-experiments,koo2010efficient,ma-zhao-2012-fourth,zhang-mcdonald-2012-generalized,tchernowitz-etal-2016-effective}, they keep some between-edge information, but require more decoding time.

In this paper, we apply first-order models, specifically the MST algorithm, and show that it is possible to keep correlations between edges without increasing the time complexity by recursively conditioning each edge score on a previous prediction of the complete dependency graph.

%% file: model.tex
\section{RNG Transformer}
\label{sec:rngtr}

\begin{figure}
  \centering
  \includegraphics[width=\linewidth]{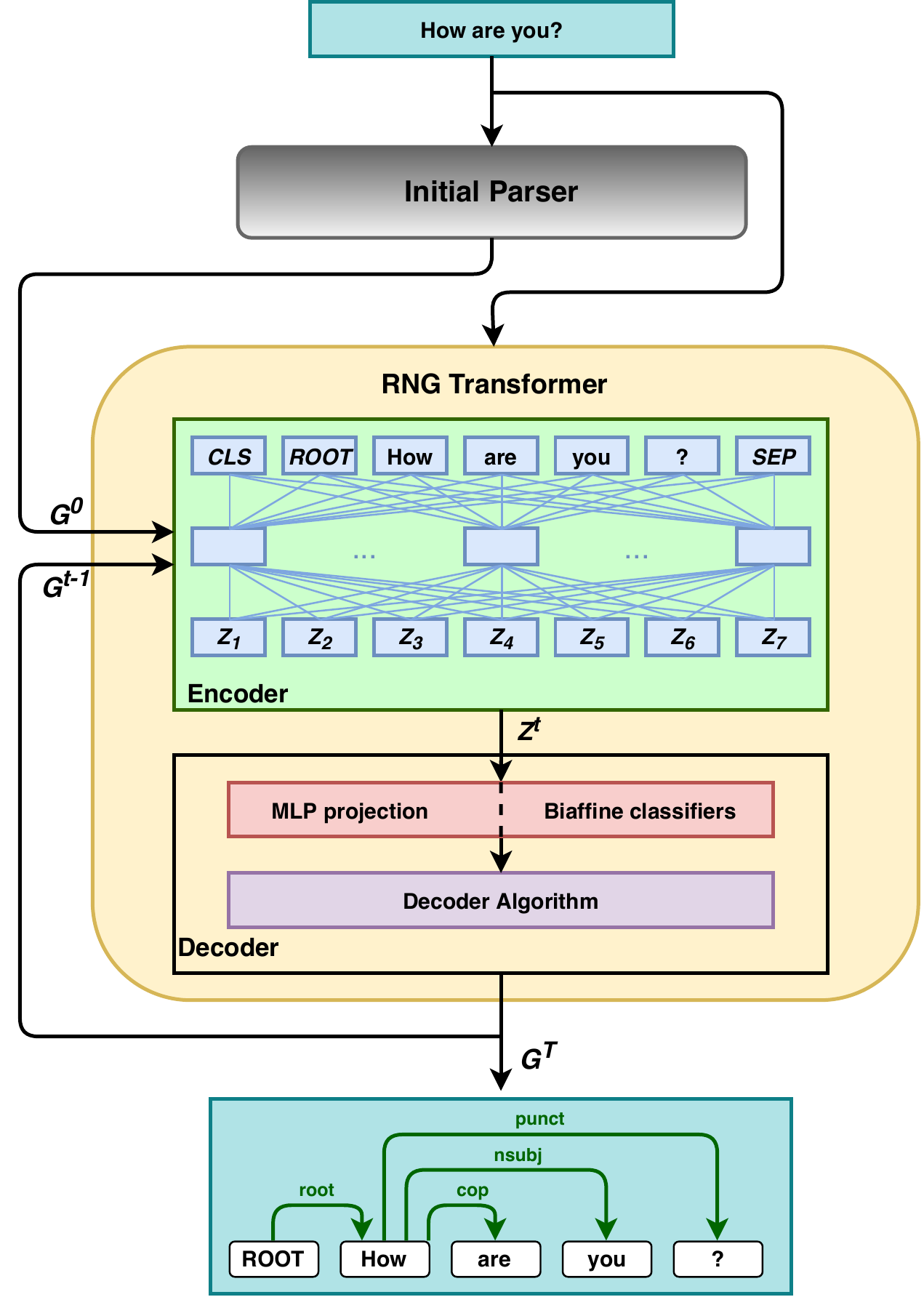}
  \caption{The Recursive Non-autoregressive Graph-to-Graph Transformer architecture.}
  \label{fig:rg2g_model}
\end{figure}

The RNG Transformer architecture is illustrated in Figure~\ref{fig:rg2g_model}, in this case, applied to dependency parsing.  The input to a RNGTr model specifies the input nodes $W = (w_1,w_2,\ldots,w_N)$ (e.g.\ a sentence), and the output is the final graph $G^T$ (e.g.\ a parse tree) over this set of nodes.  The first step is to compute an initial graph of $G^0$ over $W$, which can be done with any model.  Then each recursive iteration takes the previous graph $G^{t-1}$ as input and predicts a new graph $G^t$.

The RNGTr model predicts $G^t$ with a novel version of a Graph-to-Graph Transformer \cite{mohammadshahi2019graphtograph}.  Unlike in the work of \newcite{mohammadshahi2019graphtograph}, this G2GTr model predicts every edge of the graph in a single non-autoregressive step.  As previously, the G2GTr first encodes the input graph $G^{t-1}$ in a set of contextualised vector representations $Z = (z_1,z_2,\ldots,z_N)$, with one vector for each node of the graph.  
The decoder component then predicts the output graph $G^t$ by first computing scores for each possible edge between each pair of nodes and then applying a decoding algorithm to output the highest-scoring complete graph.  

The RNGTr model can be formalised in terms of an encoder $\operatorname{E^{RNG}}$ and a decoder $\operatorname{D^{RNG}}$:
\begin{equation}
    \begin{cases}
        Z^t = \operatorname{ E^{RNG}}(W,P,G^{t-1}) \\
        G^t = \operatorname{ D^{RNG}}(Z^t) 
    \end{cases}
    t = 1,\ldots,T
\label{eq:rng-main}
\end{equation}
where $W = (w_1,w_2,,\ldots,w_N)$ is the input sequence of tokens, $P = (p_1,p_2,,\ldots,p_N)$ is their associated properties, and $T$ is the number of refinement iterations.

In the case of dependency parsing, $W$ are the words and symbols, $P$ are their part-of-speech tags, and the predicted graph at iteration $t$ is specified as:

\begin{equation}
  \begin{split}
    & G^t = \{(i,j,l),~ 
    j = 3,\ldots,N{-}1\} \\
      &\text{where~} 2\leq i\leq N{-}1,~ l \in L
  \end{split}
\label{eq:graphnotation}
\end{equation}
Each word $w_j$ has one head (parent) $w_i$ with dependency label $l$ from the label set $L$, where the parent can also be the {\tt ROOT} symbol $w_2$ (see Section~\ref{sec:embedding}).

The following sections describe in more detail each element of the proposed RNGTr dependency parsing model.

\subsection{Encoder}
To compute the embeddings $Z^t$ for the nodes of the graph, we use the Graph-to-Graph Transformer architecture proposed by \newcite{mohammadshahi2019graphtograph}, including similar mechanism to input the previously predicted dependency graph $G^{t-1}$ to the attention mechanism.  This graph input allows the node embeddings to include both token-level and relation-level information.

\subsubsection{Input Embeddings}
\label{sec:embedding}

The RNGTr model receives a sequence of input tokens ($W$) with their associated properties ($P$) and builds a sequence of input embeddings ($X$).  For compatibility with BERT's input token representation~\cite{devlin2018bert}, the sequence of input tokens starts with {\tt CLS} and ends with {\tt SEP} symbols.  For dependency parsing, it also adds the {\tt ROOT} symbol to the front of the sentence to represent the root of the dependency tree.
To build token representation for a sequence of input tokens, we sum several vectors.  For the input words and symbols, we sum the token embeddings of a pre-trained BERT model $\operatorname{EMB}(w_i)$, and learned representations $\operatorname{EMB}(p_i)$ of their Part-of-Speech tags $p_i$.
To keep the order information of the initial sequence, we add the position embeddings of pre-trained BERT $F_i$ to our token embeddings.
The final input representations are the sum of the position embeddings and the token embeddings:
\vspace{-0.5ex}
\begin{align}
\label{eq:final_token_embedding}
\begin{split}
  x_i = F_i + \operatorname{EMB}(w_i) + \operatorname{EMB}(p_i),~ i=1,2,...,N
\end{split}
\vspace{-1ex}
\end{align}

\subsubsection{Self-Attention Mechanism}
\label{rec-attention}

\begin{figure}
  \centering
  \includegraphics[width=0.7\linewidth]{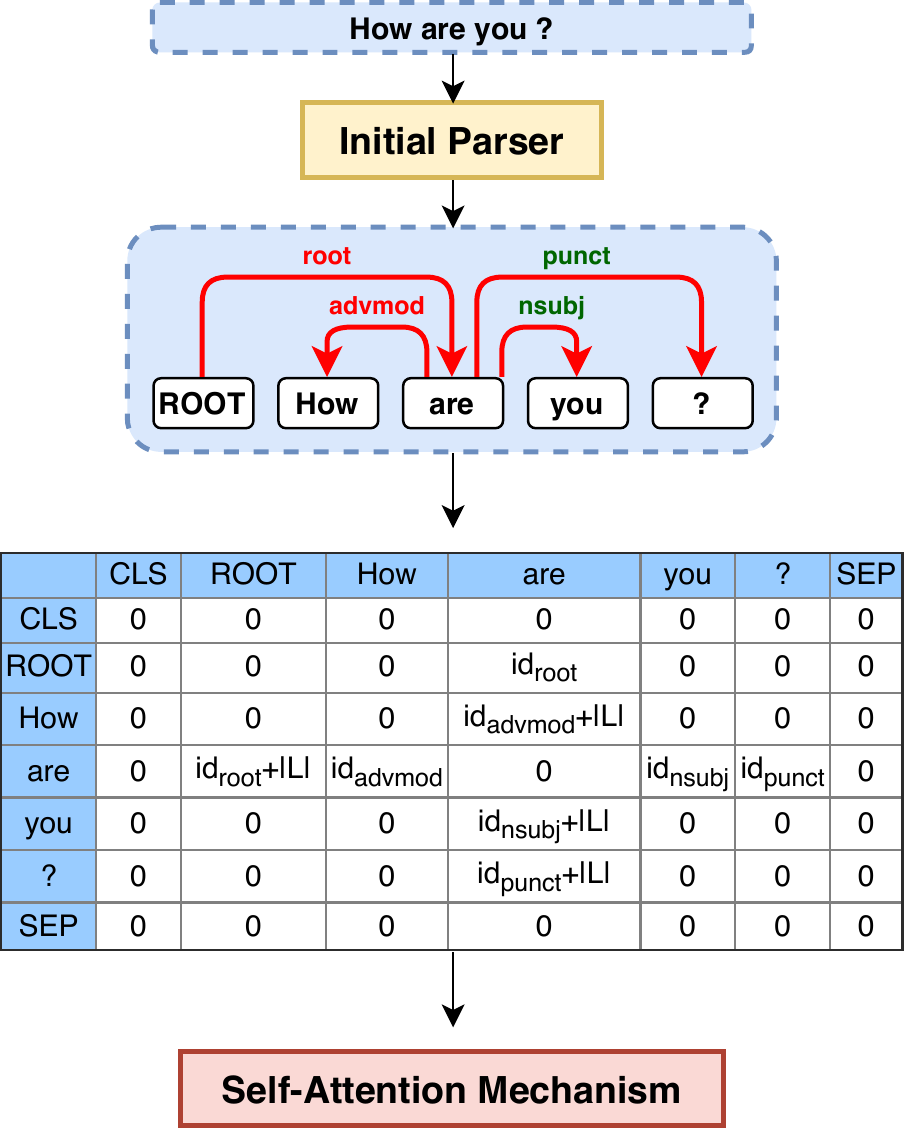}
  \caption{Example of inputting dependency graph to the self-attention mechanism.} 
  \label{fig:attention_ex}
\end{figure}

Conditioning on the previously predicted output graph $G^{t-1}$ is made possible by inputting relation embeddings to the self-attention mechanism.  This edge input method was initially proposed by \newcite{shaw-etal-2018-self} for relative position encoding, and extending to unlabelled dependency graphs in the Graph-to-Graph Transformer architecture of \newcite{mohammadshahi2019graphtograph}.  We use it to input labelled dependency graphs, by adding relation label embeddings to both the value function and the attention weight function.

Transformers have multiple layers of self-attention, each with multiple heads.  The RNGTr architecture uses the same architecture as BERT~\cite{devlin2018bert} but changes the functions used by each attention head.  Given the token embeddings $X$ at the previous layer and the input graph $G^{t-1}$, the values $A{=}(a_1,\ldots,a_N)$ computed by an attention head are: 
\begin{align}
\label{eq:g2g-attn1}
\begin{split}
a_i = \sum_j\alpha_{ij}(x_j\boldsymbol{W^V} + r^{t-1}_{ij}\boldsymbol{W^L_2})
\end{split}
\end{align}
where $r^{t-1}_{ij}$ is a one-hot vector that represents the labelled dependency relation between $i$ and $j$ in the graph $G^{t-1}$.
As shown in the matrix in Figure~\ref{fig:attention_ex},
each $r^{t-1}_{ij}$ specifies both the label and the direction of the relation ($id_{\mbox{\scriptsize\em label}}$ for $i\rightarrow j$ versus $id_{\mbox{\scriptsize\em label}}+|L|$ for $i\leftarrow j$, where $|L|$ is the number of dependency labels), or specifies {\sc none} (as $0$).  $\boldsymbol{W^L_2} \in R^{(2|L|+1) \times d}$ are the learned relation embeddings.
The attention weights $\alpha_{ij}$ are a Softmax applied to the attention function:
\begin{align}
\nonumber
\alpha_{ij} =~& \frac{\operatorname{exp}(e_{ij})}{\sum \operatorname{exp}(e_{ij})}
\\
\label{eq:g2g-attn3}
e_{ij} =~& \frac{(x_i\boldsymbol{W^Q})(x_j\boldsymbol{W^K}+\operatorname{LN}(r^{t-1}_{ij}\boldsymbol{W^L_1}))}{\sqrt{d}}
\end{align}
where $\boldsymbol{W^L_1} \in R^{(2|L|+1) \times d}$ are different learned relation embeddings. $\operatorname{LN(\cdot})$ is the layer normalisation function, used for better convergence.

Equations~\eqref{eq:g2g-attn1} and~\eqref{eq:g2g-attn3} constitute the mechanism by which each iteration of refinement can condition on the previous graph.  Instead of the more common approach of hard-coding some attention heads to represent a relation (e.g.\ \newcite{ji-etal-2019-graph}), all attention heads can learn for themselves how to use the information about relations.

\subsection{Decoder}
\label{sec:decoder}

The decoder uses the token embeddings $Z^t$ produced by the encoder to predict the new graph $G^t$.  It consists of two components, a scoring function, and a decoding algorithm.  The graph found by the decoding algorithm is the output graph $G^t$ of the decoder.
Here we propose components for dependency parsing.

\subsubsection{Scoring Function}
\label{sec:biaffine}

We first produce four distinct vectors for each token embedding $z^t_i$ from the encoder by passing it through four feed-forward layers.
\vspace{-0.5ex}
\begin{align}
\label{eq:mlp_output}
\begin{split}
  &z_i^{t,(arc-dep)} = \operatorname{MLP}^{(arc-dep)}(z^t_i) \\
  &z_i^{t,(arc-head)} = \operatorname{MLP}^{(arc-head)}(z^t_i) \\
  &z_i^{t,(rel-dep)} = \operatorname{MLP}^{(rel-dep)}(z^t_i) \\
  &z_i^{t,(rel-head)} = \operatorname{MLP}^{(rel-head)}(z^t_i) 
\end{split}
\vspace{-1ex}
\end{align} 
where the $\operatorname{MLP}$'s are all one-layer feed-forward networks with $\operatorname{LeakyReLU}$ activation functions.

These token embeddings are used to compute probabilities for every possible dependency relation, both unlabelled and labelled, similarly to \newcite{dozat2016deep}.  The distribution of the unlabelled dependency graph is estimated using, for each token $i$, a Biaffine classifier over possible heads $j$ applied to $z_i^{t,(arc-dep)}$ and $z_j^{t,(arc-head)}$.  Then for each pair $i,j$, the distribution over labels given an unlabelled dependency relation is estimated using a Biaffine classifier applied to $z_i^{t,(rel-dep)}$ and $z_j^{t,(rel-head)}$.

\subsubsection{Decoding Algorithms}
\label{sec:decode_alg}

The scoring function estimates a distribution over graphs, but the RNGTr architecture requires the decoder to output a single graph $G^t$.  Choosing this graph is complicated by the fact that the scoring function is non-autoregressive. Thus the estimate consists of multiple independent components, and thus there is no guarantee that every graph in this distribution is a valid dependency graph.

We take two approaches to this problem, one for intermediate parses $G^t$ and one for the final dependency parse $G^T$.
To speed up each refinement iteration, we ignore this problem for intermediate dependency graphs.
We build these graphs by simply applying $\operatorname{argmax}$ independently to find the head of each node.  This may result in graphs with loops, which are not trees, but this does not seem to cause problems for later refinement iterations.\footnote{We leave to future work the investigation of different decoding strategies that keep both speed and well-formedness for the intermediate predicted graphs.}
For the final output dependency tree, we use the maximum spanning tree algorithm, specifically the Chu-Liu/Edmonds algorithm~\cite{chi-1999-statistical,edmonds1967optimum}, to find the highest scoring valid dependency tree.  This is necessary to avoid problems when running the evaluation scripts.
The asymptotic complexity of the full model is determined by the complexity of this algorithm.\footnote{The Tarjan variation~\cite{karger1995randomized} of Chu-Liu/Edmonds algorithm computes the highest-scoring tree in $O(n^2)$ for dense graphs, which is the case here.}

\subsection{Training}
\label{sec:training}

The RNG Transformer model is trained separately on each refinement iteration.  Standard gradient descent techniques are used, with cross-entropy loss for each edge prediction.  Error is not backpropagated across iterations of refinement, because no continuous values are being passed from one iteration to another, only a discrete dependency tree.

\paragraph{Stopping Criterion:}
\label{sec:stopping}
In the RNG Transformer architecture, the refinement of the predicted graph can be done an arbitrary number of times, since the same encoder and decoder parameters are used at each iteration.  In the experiments below, we place a limit on the maximum number of iterations.  But sometimes the model converges to an output graph before this limit is reached, simply copying this graph during later iterations. During training, to avoid multiple iterations where the model is trained to simply copy the input graph, the refinement iterations are stopped if the new predicted dependency graph is the same as the input graph.  At test time, we also stop computation in this case, but the output of the model is not affected.

\section{Initial Parsers}
\label{sec:initial}

The RNGTr architecture requires a graph $G^0$ to initialise the iterative refinement.  We consider several initial parsers to produce this graph.  
To leverage previous work on dependency parsing and provide a controlled comparison to the state-of-the-art, we use parsing models from the recent literature as both baselines and initial parsers.  To evaluate the importance of the initial parse, we also consider a setting where the initial parse is empty, so the first complete dependency tree is predicted by the RNGTr model itself.  Finally, the success of our RNGTr dependency parsing model leads us to propose an initial parsing model with the same design, so that we can control for the parser design in measuring the importance of the RNG Transformer's iterative refinement. 

\paragraph{SynTr model}
We call this initial parser the Syntactic Transformer (SynTr) model.
It is the same as one iteration of the RNGTr model shown in Figure~\ref{fig:rg2g_model} and defined in Section~\ref{sec:rngtr}, except that there is no graph input to the encoder.  
Analogously to \eqref{eq:rng-main}, $G^0$ is computed as:
\begin{equation}
    \begin{cases}
        Z^0 = \operatorname{E^{SYNTR}}(W,P) \\
        G^0 = \operatorname{ D^{SYNTR}}(Z^0)
    \end{cases}  
\label{eq:syntr}
\end{equation}
where $\operatorname{E^{SYNTR}}$ and $\operatorname{D^{SYNTR}}$ are the SynTr encoder and decoder, respectively.
For the encoder, we use the Transformer architecture of BERT~\cite{devlin2018bert} and initialise with pre-trained parameters of BERT.  The token embeddings of the final layer are used for $Z^0$. 
For the decoder, we use the same scoring function as described in Section~\ref{sec:decoder}, and apply Chu-Liu/Edmonds decoding algorithm~\cite{chi-1999-statistical,edmonds1967optimum} to find the highest scoring tree.

This SynTr parsing model is very similar to the UDify parsing model proposed by \newcite{Kondratyuk_2019}.  One difference which seems to be important for the results reported in Section~\ref{sec:ud-results} is in the way BERT token segmentation is handled.  When BERT segments a word into sub-words, UDify seems only to encode the first segment, whereas SynTr encodes all segments and only decodes with the first segment, as discussed in Section~\ref{sec:implement}.  Also, UDify decodes with an attention-based mixture of encoder layers, whereas SynTr only uses the last layer.

%% file: implement.tex
\section{Experimental Setup}

\subsection{Datasets}

To evaluate our models, we apply them on several kinds of datasets, namely Universal Dependency (UD) Treebanks, Penn Treebanks, and the German CoNLL 2009 Treebank. 
For our evaluation on Universal Dependency Treebanks (UD v2.3)~\cite{11234/1-2895}, we select languages based on the criteria proposed in \newcite{Lhoneux2017OldSV}, and adapted by \newcite{smith-etal-2018-investigation}. This set contains several languages with different language families, scripts, character set sizes, morphological complexity, and training sizes and domains. 
For our evaluation of Penn Treebanks, we use the English and Chinese Penn Treebanks~\cite{marcus-etal-1993-building,xue-etal-2002-building}. For English, we use the same setting as defined in \newcite{mohammadshahi2019graphtograph}.
For Chinese, we apply the same setup as described in \newcite{chen-manning-2014-fast},
including the use of gold PoS tags. 
For our evaluation on the German Treebank of the CoNLL 2009 shared task~\cite{hajic-etal-2009-conll}, we apply the same setup as defined in \newcite{kuncoro-etal-2016-distilling}.
Following \newcite{hajic-etal-2009-conll,11234/1-2895}, we keep punctuation for evaluation on the UD Treebanks and the German corpus and remove it for the Penn Treebanks~\cite{nilsson-nivre-2008-malteval}.

\subsection{Baseline Models}

For UD Treebanks, we compare to several baseline parsing models.
We use the monolingual parser proposed by \newcite{Kulmizev_2019}, which
uses BERT~\cite{devlin2018bert} and ELMo~\cite{peters-etal-2018-deep} embeddings as additional input features. In addition, we compare to the multilingual multi-task models proposed by \newcite{Kondratyuk_2019} and \newcite{straka-2018-udpipe}. UDify~\cite{Kondratyuk_2019} is a multilingual multi-task model.
UDPipe~\cite{straka-2018-udpipe} is one of the winners of CoNLL 2018 Shared Task~\cite{zeman-etal-2018-conll}.
For a fair comparison, we report the scores of UDPipe from \newcite{Kondratyuk_2019} using gold segmentation.
UDify is on average the best performing of these baseline models, so we use it as one of our initial parsers in the RNGTr model.

For Penn Treebanks and the German CoNLL 2009 corpus, we compare our models with previous state-of-the-art transition-based, and graph-based models, including the Biaffine parser~\cite{dozat2016deep}, which includes the same decoder as our model.
We also use the Biaffine parser as an initial parser for the RNGTr model.

\subsection{Implementation Details}
\label{sec:implement}

The encoder is initialised with pre-trained BERT~\cite{devlin2018bert} models with 12 self-attention layers.
All hyper-parameters are provided in Appendix~\ref{apx:hyper-param}.

Since the wordpiece tokeniser~\cite{wu2016google} of BERT differs from that used in the dependency corpora, we apply the BERT tokeniser to each corpus word and input all the resulting sub-words to the encoder.  For the input of dependency relations, each dependency between two words is specified as a relationship between their first sub-words.  We also input a new relationship between each non-first sub-word and its associated first sub-word as its head.
For the prediction of dependency relations, only the encoder embedding of the first sub-word of each word is used by the decoder.\footnote{ In preliminary experiments, we found that predicting dependencies using the first sub-words achieves better or similar results compared to using the last sub-word or all sub-words of each word.}  The decoder predicts each dependency as a relation between the first sub-words of the corresponding words.  Finally, for proper evaluation, we map the predicted sub-word heads and dependents to their original word positions in the corpus.

%% file: results.tex
\begin{table}[tb]
\centering
  \begin{adjustbox}{width=\linewidth}
    \begin{tabular}{|l|c|c|}
    \hline
    Model & UAS & LAS\\
    \hline
     SynTr & 75.62 & 70.04 \\
     SynTr+RNGTr (T=1) & 76.37 & 70.67 \\
     SynTr+RNGTr (T=3) w/o \textit{stop} & 76.33 & 70.61  \\
     SynTr+RNGTr (T=3) & 76.29 & 70.84  \\ 
    \hline
     UDify~\cite{Kondratyuk_2019} & 72.78 & 65.48 \\
     UDify+RNGTr (T=1) & 74.13 & 68.60 \\
     UDify+RNGTr (T=3) w/o \textit{stop} & 75.68 & 70.32 \\
     UDify+RNGTr (T=3) & 75.91 & 70.66 \\
    \hline
    
    \end{tabular}
  \end{adjustbox}
\caption{\label{result-turkish} Dependency parsing scores for different variations of the RNG Transformer model on the development set of UD Turkish Treebank (IMST).}
\end{table}

\begin{table*}[bt]
  \centering
  \begin{adjustbox}{width=\linewidth}
  \begin{tabular}{|c|c|c|c|cl|cl|c|}
    \hline
    \multirow{2}{*}{Language} & Train &  
    Mono & Multi & Multi & Multi+Mono & Mono & Mono & Mono \\
    &  Size & [1] & UDPipe & UDify & UDify+RNGTr & SynTr & SynTr+RNGTr & Empty+RNGTr \\
    \hline
    Arabic & 6.1K & 81.8 & 82.94 & 82.88 & \textbf{85.93} (+17.81\%) & \textbf{86.23} & \textbf{86.31} (+0.58\%) & \textbf{86.05} \\
    Basque & 5.4K & 79.8 & 82.86 & 80.97 & 87.55 (+34.57\%) & 87.49 & \textbf{88.2} (+5.68\%) & \textbf{87.96} \\
    Chinese & 4K & 83.4 & 80.5 & 83.75 & 89.05 (+32.62\%) & 89.53 & \textbf{90.48} (+9.08\%)  & 89.82 \\
    English & 12.5K & 87.6 & 86.97 & 88.5 & 91.23 (+23.74\%) & \textbf{91.41} & \textbf{91.52} (+1.28\%) & 91.23 \\
    Finnish & 12.2K & 83.9 & 87.46 & 82.03 & \textbf{91.87} (+54.76\%) & \textbf{91.80} & \textbf{91.92} (+1.46\%) & \textbf{91.78} \\
    Hebrew & 5.2K & 85.9 & 86.86 & 88.11 & 90.80 (+22.62\%) & \textbf{91.07} & \textbf{91.32} (+2.79\%) & 90.56 \\
    Hindi & 13.3K & 90.8 & 91.83 & 91.46 & 93.94 (+29.04\%) & 93.95 & \textbf{94.21} (+4.3\%) & 93.97 \\
    Italian &13.1K & 91.7 & 91.54 & 93.69 & 94.65 (+15.21\%) & \textbf{95.08} & \textbf{95.16} (+1.62\%) & \textbf{94.96} \\
    Japanese & 7.1K & 92.1 & 93.73 & 92.08 & \textbf{95.41} (+42.06\%) & \textbf{95.66} & \textbf{95.71} (+1.16\%) & \textbf{95.56} \\
    Korean & 4.4K & 84.2 & 84.24 & 74.26 & \textbf{89.12} (+57.73\%) & \textbf{89.29} & \textbf{89.45} (+1.5\%) & \textbf{89.1} \\
    Russian & 48.8K & 91.0 & 92.32 & 93.13 & \textbf{94.51} (+20.09\%) & \textbf{94.60} & \textbf{94.47} (-2.4\%) & 94.31 \\
    Swedish & 4.3K & 86.9 & 86.61 & 89.03 & 92.02 (+27.26\%) & 92.03 & \textbf{92.46} (+5.4\%) & \textbf{92.40} \\
    Turkish & 3.7K & 64.9 & 67.56 & 67.44 & 72.07 (+14.22\%) & \textbf{72.52} & \textbf{73.08} (+2.04\%) & 71.99 \\
    \hline
    Average & - & 84.9 & 85.81 & 85.18 & 89.86 & 90.05 & 90.33 & 89.98 \\
    \hline
  \end{tabular}
  \end{adjustbox}
  \caption{\label{ud-las} Labelled attachment scores on UD Treebanks for monolingual ([1]~\cite{Kulmizev_2019} and SynTr) and multilingual (UDPipe~\cite{straka-2018-udpipe} and UDify~\cite{Kondratyuk_2019}) baselines, and the refined models (+RNGTr) pre-trained with BERT~\cite{devlin2018bert}.
    The relative error reduction from RNGTr refinement is shown in parentheses.
    Bold scores are not significantly different from the best score in that row (with $\alpha=0.01$).
  }
\end{table*}

\begin{table*}[tb]
  \centering
  \begin{adjustbox}{width=\textwidth}
  \begin{tabular}{|c|c|cc|cc|cc|}
    \hline
    \multirow{2}{*}{Model} &
      &
      \multicolumn{2}{c|}{English\,(PTB)} &
      \multicolumn{2}{c|}{Chinese\,(CTB)} & 
      \multicolumn{2}{c|}{German\,(CoNLL)} \\
    & Type & UAS & LAS & UAS & LAS & UAS & LAS \\
    \hline
    \newcite{chen-manning-2014-fast} & T & 91.8 & 89.6 & 83.9 & 82.4 & - & - \\
    \newcite{dyer-etal-2015-transition} & T & 93.1 & 90.9 & 87.2 & 85.7 & - & - \\
    \newcite{ballesteros-etal-2016-training} & T & 93.56 & 91.42 & 87.65 & 86.21 & 88.83 & 86.10 \\
    \newcite{cross-huang-2016-incremental} & T & 93.42 & 91.36 & 86.35 & 85.71 & - & - \\
    \newcite{weiss-etal-2015-structured} & T & 94.26 & 92.41 & - & - & - & - \\
    \newcite{andor-etal-2016-globally} & T & 94.61 & 92.79 & - & - & 90.91 & 89.15 \\
    \newcite{mohammadshahi2019graphtograph} & T & 96.11 & 94.33 & - & - & - & - \\
    \newcite{ma-etal-2018-stack} & T & 95.87 & 94.19 & 90.59 & 89.29 & 93.65 & 92.11 \\
    \newcite{FerGomNAACL2019} & T & 96.04 & 94.43 & - & - & - & - \\
    \hline
    \newcite{kiperwasser-goldberg-2016-simple} & G & 93.1 & 91.0 & 86.6 & 85.1 & - & - \\
    \newcite{wang-chang-2016-graph} & G & 94.08 & 91.82 & 87.55 & 86.23 & - & - \\
    \newcite{cheng-etal-2016-bi} & G & 94.10 & 91.49 & 88.1 & 85.7 & - & - \\
    \newcite{kuncoro-etal-2016-distilling} & G & 94.26 & 92.06 & 88.87 & 87.30 & 91.60 & 89.24 \\
    \newcite{ma2017neural} & G & 94.88 & 92.98 & 89.05 & 87.74 & 92.58 & 90.54 \\
    \newcite{ji-etal-2019-graph} & G & 95.97 & 94.31 & - & -& - & - \\
    \hline
    \newcite{li2019global}+ELMo & G & 96.37 & 94.57 & 90.51 & 89.45 & - & - \\
    \newcite{li2019global}+BERT & G & 96.44 & 94.63 & 90.89 & 89.73 & - & - \\
    \hline
    Biaffine~\cite{dozat2016deep} & G & 95.74 & 94.08 & 89.30 & 88.23 & 93.46 & 91.44 \\
    Biaffine+RNGTr & G & 96.44 & 94.71 & 91.85 & 90.12 & 94.68 & 93.30 \\
    \hline
    SynTr & G & \textbf{96.60} & \textbf{94.94} & 92.42 & 90.67 & \textbf{95.11} & \textbf{93.98} \\
    SynTr+RNGTr & G & \textbf{96.66} & \textbf{95.01} & \textbf{92.98} & \textbf{91.18} & \textbf{95.28} & \textbf{94.02} \\
    \hline
  \end{tabular}
  \end{adjustbox}
  \caption{\label{penn-results} Comparison of our models to previous SOTA models on English (PTB) and Chinese (CTB5.1) Penn Treebanks, and German CoNLL 2009 shared task treebank. "T" and "G" specify "Transition-based" and "Graph-based" models.
    Bold scores are not significantly different from the best score in that column (with $\alpha=0.01$).
  }
\end{table*}

\section{Results and Discussion}

After some initial experiments to determine the maximum number of refinement iterations, we report the performance of the RNG Transformer model on the UD treebanks, Penn treebanks, and German CoNLL 2009 treebank.\footnote{The number of parameters and run times of each model on the UD and Penn Treebanks are provided in Appendix~\ref{apx:runtime}.}
The RNGTr models perform substantially better than previously proposed models on every dataset, and RNGTr refinement improves over its initial parser for almost every dataset. 
We also perform various analyses to understand these results better.

\subsection{The Number of Refinement Iterations}
\label{sec:iteration-results}

Before conducting a large number of experiments, we investigate how many iterations of refinement are useful, given the computational costs of additional iterations.  
We evaluate different variations of our RNG Transformer model on the Turkish Treebank (Table~\ref{result-turkish}).\footnote{We choose the Turkish Treebank because it is a low-resource Treebank and there are more errors in the initial parse for RNGTr to correct.} We use both SynTr and UDify as initial parsers. The SynTr model significantly outperforms the UDify model, so the errors are harder to correct by adding the RNGTr model (2.67\% for SynTr versus 15.01\% for UDify of relative error reduction in LAS after integration).  In both cases, three iterations of refinement achieve more improvement than one iteration, but not by a large enough margin to suggest the need for additional iterations.  The further analysis reported in Section~\ref{sec:refine} supports the conclusion that, in general, additional iteration would neither help nor hurt accuracy.  The results in Table~\ref{result-turkish} also show that it is better to include the stopping strategy described in Section~\ref{sec:stopping}.
In subsequent experiments, we use three refinement iterations with the stopping strategy, unless mentioned otherwise.

\subsection{UD Treebank Results}
\label{sec:ud-results}

Results for the UD treebanks are reported in Table~\ref{ud-las}.
We compare our models with previous state-of-the-art results (both trained mono-lingually and multi-lingually), based on labelled attachment score.\footnote{Unlabelled attachment scores are provided in Appendix~\ref{apx:uas-ud}.  All results are computed with the official CoNLL 2018 shared task evaluation script~(\url{https://universaldependencies.org/conll18/evaluation.html}).}

The results with RNGTr refinement demonstrate the effectiveness of the RNGTr model at refining an initial dependency graph.
First, the UDify+RNGTr model achieves significantly better LAS performance than the UDify model in all languages.
Second, although 
the SynTr model significantly outperforms previous state-of-the-art models on all these UD Treebanks,\footnote{ In particular, SynTr significantly outperforms UDify, even though they are very similar models. In addition to the model differences discussed in Section~\ref{sec:initial},
there are some differences in the way UDify and SynTr models are trained that might explain this improvement, in particular, that UDify is a multi-lingual multi-task model, whereas
SynTr is a mono-lingual single-task model.
}
the SynTr+RNGTr model achieves further significant improvement over SynTr in four languages, and no significant degradation in any language.  Of the nine languages where there is no significant difference between SynTr and SynTr+RNGTr for the given test sets, RNGTr refinement results in higher LAS in eight languages and lower LAS in only one (Russian).

The improvement of SynTr+RNGTr over SynTr is particularly interesting because it is a controlled demonstration of the effectiveness of the graph refinement method of RNGTr.  The only difference between the SynTr model and the final iteration of the SynTr+RNGTr model is the graph inputs from the previous iteration (Equations~\eqref{eq:syntr} versus \eqref{eq:rng-main}).
By conditioning on the full dependency graph, the SynTr+RNGTr model's final RNGTr iteration can capture any kind of correlation in the dependency graph, including both global and between-edge correlations both locally and over long distances.
This result also further demonstrates the generality and effectiveness of the G2GTr architecture for conditioning on graphs (Equations~\eqref{eq:g2g-attn1} and~\eqref{eq:g2g-attn3}).

As expected, we get more improvement when combining the RNGTr model with UDify, because UDify's initial dependency graph contains more incorrect dependency relations for RNGTr to correct.
But after refinement, there is surprisingly little difference between the performance of the UDify+RNGTr and SynTr+RNGTr models, suggesting that RNGTr is powerful enough to correct any initial parse.  
To investigate the power of the RNGTr architecture to correct any initial parse, we also show results for a model with an empty initial parse, Empty+RNGTr.  For this model, we run four iterations of refinement (T=4), so that the amount of computation is the same as for SynTr+RNGTr.
The Empty+RNGTr model achieves competitive results with the UDify+RNGTr model (i.e.\ above the previous state-of-the-art), and close to the results for SynTr+RNGTr.  This accuracy is achieved despite the fact that the Empty+RNGTr model has half as many parameters as the UDify+RNGtr model and the SynTr+RNGTr model since it has no separate initial parser.
These Empty+RNGTr results indicate that RNGTr architecture is a very powerful method for graph refinement.

\subsection{Penn Treebank and German corpus Results}

UAS and LAS results for the Penn Treebanks and German CoNLL 2009 Treebank are reported in Table~\ref{penn-results}.
We compare to the results of previous state-of-the-art models and SynTr, and we use the RNGTr model to refine both the Biaffine parser~\cite{dozat2016deep} and SynTr, on all Treebanks.\footnote{Results are calculated with the official evaluation script: (\url{https://depparse.uvt.nl/}). For German, we use \url{https://ufal.mff.cuni.cz/conll2009-st/eval-data.html}.}

\begin{figure*}[!ht]
  \centering
  \hspace{-2ex}
  \includegraphics[width=0.33\linewidth]{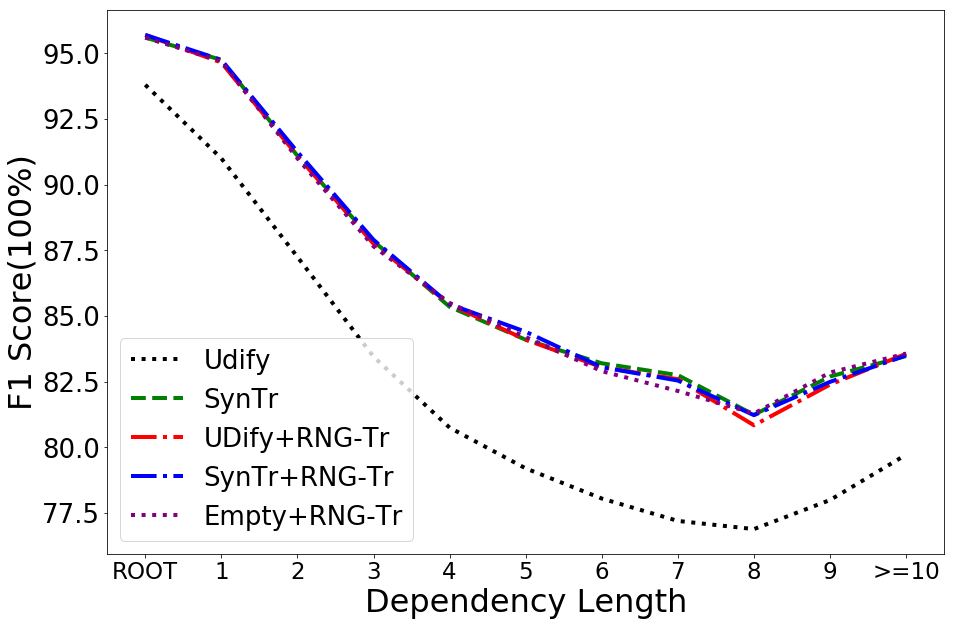}
  \includegraphics[width=0.33\linewidth]{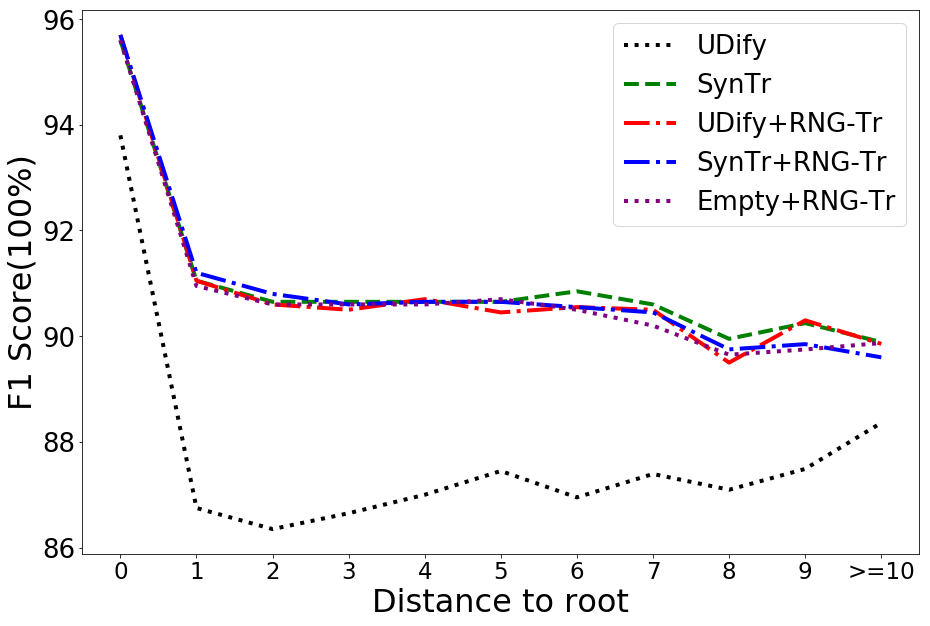}
  \includegraphics[width=0.33\linewidth]{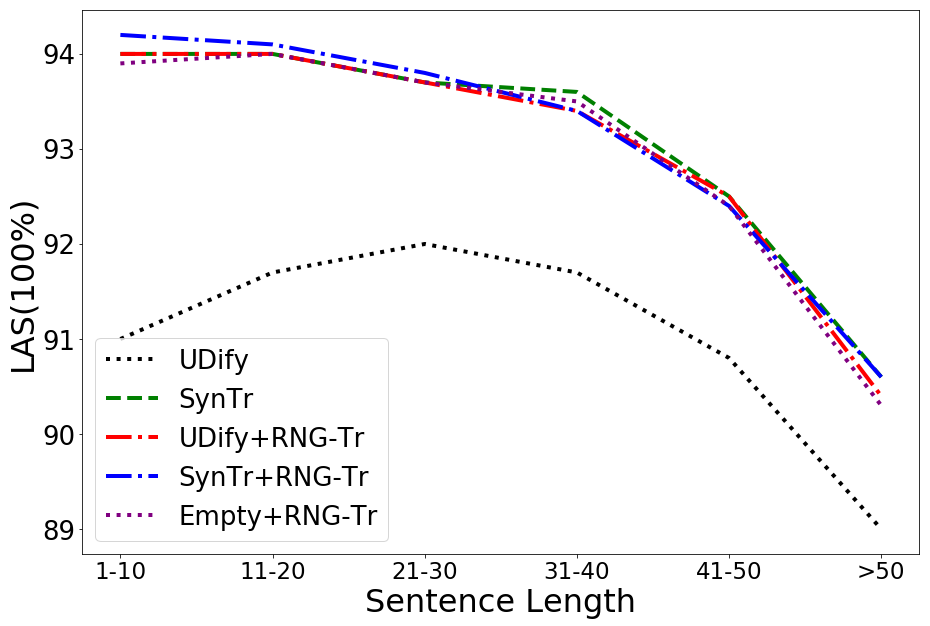}
  \hspace{-2ex}
  \caption{Error analysis, on the concatenation of UD Treebanks, of initial parsers (UDify and SynTr), their integration with the RNGTr model, and the Empty+RNGTr model.
    \vspace{-1ex}
  }
  \label{fig:errors-ud}
\end{figure*}

Again, the SynTr model significantly outperforms previous state-of-the-art models, with a 5.78\%, 9.15\%, and 23.7\% LAS relative error reduction in English, Chinese, and German, respectively.  Despite this level of accuracy, adding RNGTr refinement improves accuracy further under both UAS and LAS.  For the Chinese Treebank, this improvement is significant, with a 5.46\% LAS relative error reduction.
When RNGTr refinement is applied to the output of the Biaffine parser~\cite{dozat2016deep}, it achieves a LAS relative error reduction of 10.64\% for the English Treebank, 16.05\% for the Chinese Treebank, and 27.72\% for the German Treebank.
These improvements, even over such strong initial parsers, again demonstrate the effectiveness of the RNGTr architecture for graph refinement.

\subsection{Error Analysis}
\label{sec:error-an}

To better understand the distribution of errors for our models, we follow \newcite{mcdonald-nivre-2011-analyzing} and plot labelled attachment scores as a function of dependency length, sentence length and distance to root.\footnote{We use the MaltEval tool~\cite{nilsson-nivre-2008-malteval} for calculating accuracies in all cases.}
We compare the distributions of errors made by the UDify~\cite{Kondratyuk_2019}, SynTr, and refined models (UDify+RNGTr, SynTr+RNGTr, and Empty+RNGTr).  Figure~\ref{fig:errors-ud} shows the accuracies of the different models on the concatenation of all development sets of UD Treebanks.
Results show that applying RNGTr refinement to the UDify model results in a substantial improvement in accuracy across the full range of values in all cases,  
and little difference in the error profile between the better performing models.
In all the plots, the gains from RNGTr refinement are more pronounced for the more difficult cases, where a larger or more global view of the structure is beneficial.

As shown in the leftmost plot of Figure~\ref{fig:errors-ud}, adding RNGTr refinement to UDify results in particular gains for the longer dependencies, which are more likely to interact with other dependencies.
The middle plot illustrates the accuracy of models as a function of the distance to the root of the dependency tree, which is calculated as the number of dependency relations from the dependent to the root.
When we add RNGTr refinement to the UDify parser, we get particular gains for the problematic middle depths, which are neither the root nor leaves.  Here, SynTr+RNGTr is also particularly strong on these high nodes, whereas SynTr is particularly strong on low nodes.
In the plot by sentence length, the larger improvements from adding RNGTr refinement (both to UDify and SynTr) are for the shorter sentences, which are surprisingly difficult for UDify.  Presumably, these shorter sentences tend to be more idiosyncratic, which is better handled with a global view of the structure. (See Figure~\ref{fig:refineexample} for an example.)
In all these cases, the ability of RNGTr to capture any kind of correlation in the dependency graph gives the model a larger and more global view of the correct output structure.

\begin{figure*}[!ht]
  \centering
  \hspace{-2ex}
  \includegraphics[width=0.4\linewidth]{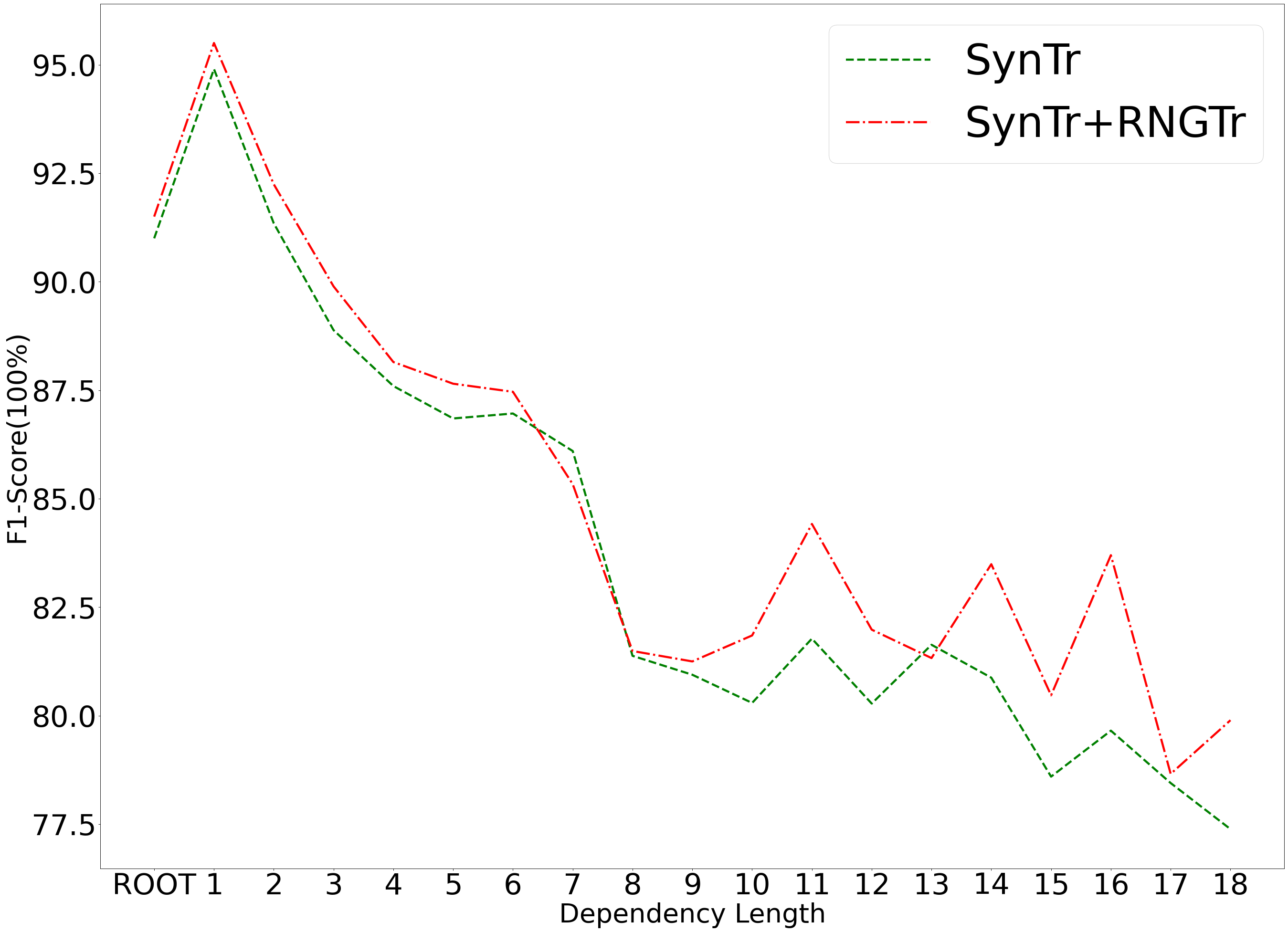}
  \includegraphics[width=0.4\linewidth]{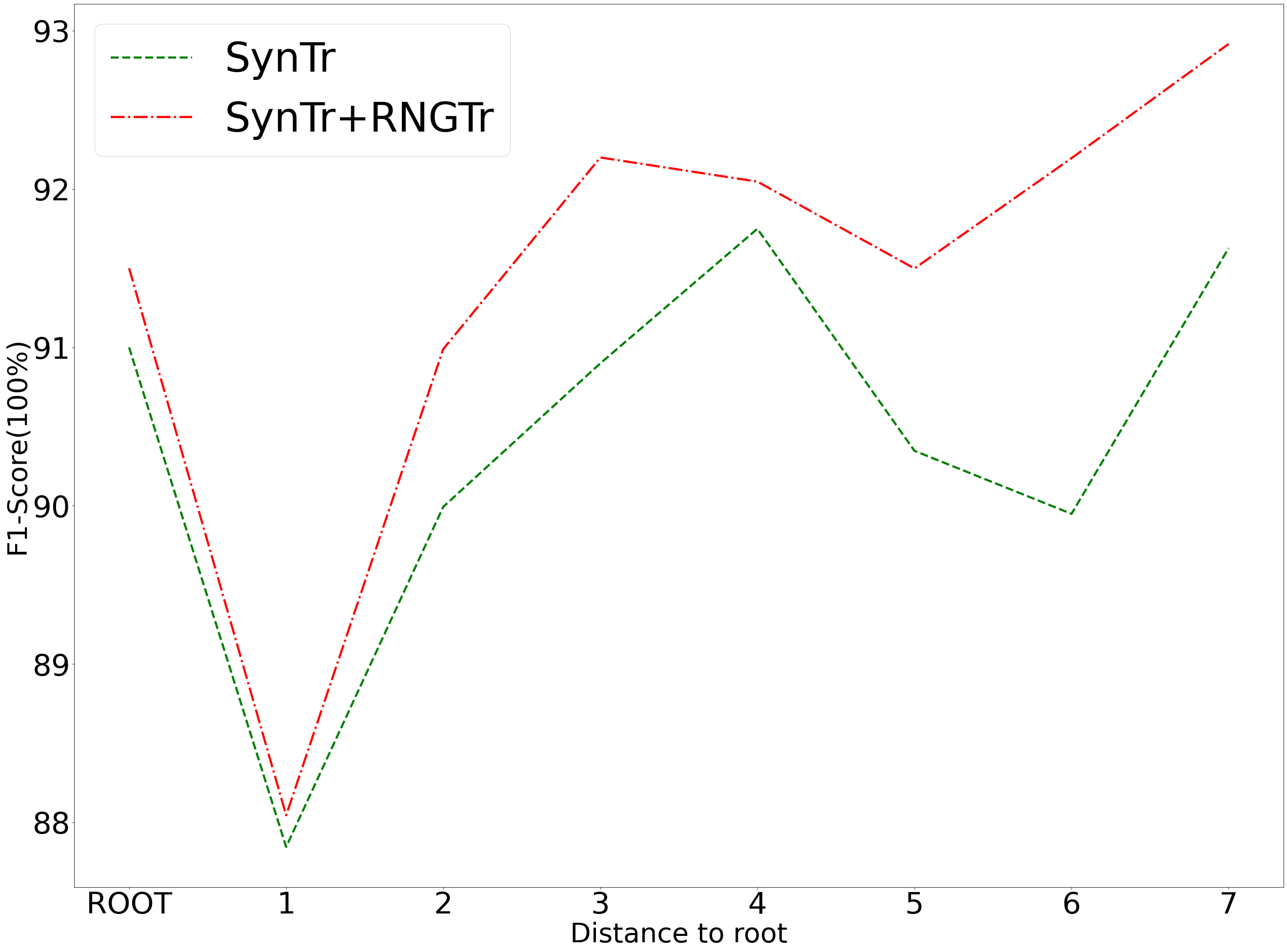}
  \hspace{-2ex}
  \caption{Error analysis of SynTr and SynTr+RNGTr models on Chinese CTB Treebank.
    \vspace{-1ex}
    \label{fig:error-ctb}
  }
\end{figure*}

To further analyse where RNGTr refinement is resulting in improvements,
we compare the error profiles of the  SynTr and SynTr+RNGTr models on the Chinese Penn Treebank, where adding RNGTr refinement to SynTr results in significant improvement (see Table~\ref{penn-results}).
As shown in Figure~\ref{fig:error-ctb}, RNGTr refinement results in particular improvement on longer dependencies (left plot), and on middle and greater depth nodes (right plot), again showing that RNGTr does particularly well on the difficult cases with more interactions with other dependencies.

\subsection{Refinement Analysis}
\label{sec:refine}

\begin{figure}
\centering
  \includegraphics[width=0.55\linewidth]{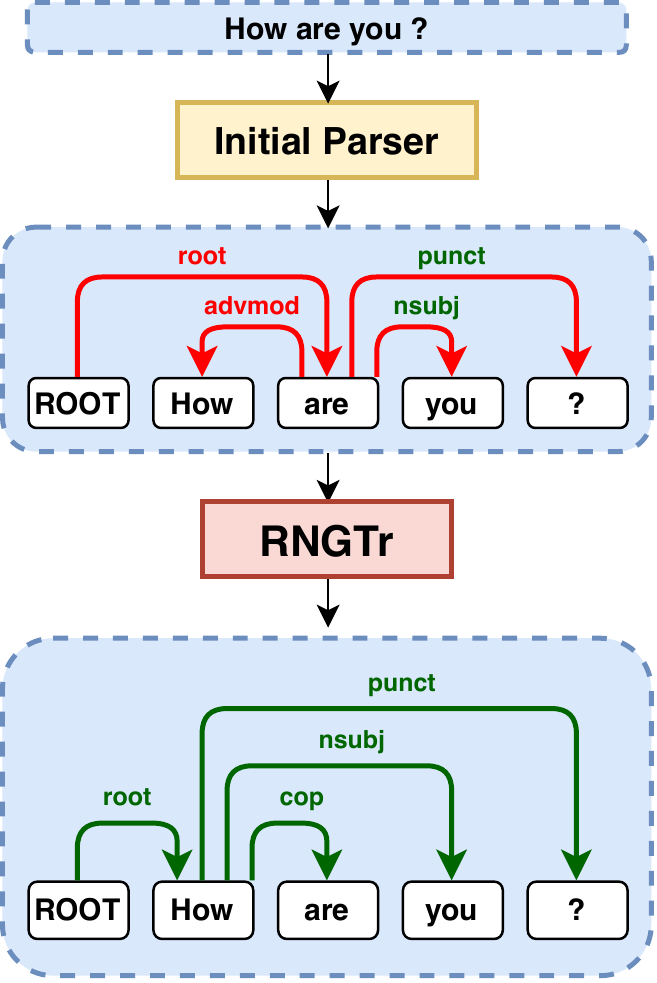}
  \caption{The shortest example corrected by UDify+RNGTr in the English UD Treebank.}
  \label{fig:refineexample}
\end{figure}

To better understand how the RNG Transformer model is doing refinement, we perform several analyses of the trained UDify+RNGTr model.\footnote{We choose UDify as the initial parser because the RNGTr model makes more changes to the parses of UDify than SynTr, so we can more easily analyse these changes. Results with SynTr as the initial parser are provided in Appendix~\ref{apx:syntr-refine}.}
An example of this refinement is shown in Figure~\ref{fig:refineexample}, where the UDify model predicts an incorrect dependency graph, but the RNGTr model modifies it to build the gold dependency tree.

\paragraph{Refinements by Iteration:}
To measure the accuracy gained from refinement at different iterations, we define the following metric:
\begin{equation}
{\tt REL}^t = {\tt RER}({\tt LAS}^{t-1},{\tt LAS}^t)
\label{eq:refine}
\end{equation}
where ${\tt RER}$ is relative error reduction, and $t$ is the refinement iteration. ${\tt LAS}^0$ is the accuracy of the initial parser, UDify in this case.

\begin{table}
\centering
  \begin{adjustbox}{width=\linewidth}
  
  \begin{tabular}{|c|c|c|c|}
    \hline
    Dataset Type & $t=1$ & $t=2$ & $t=3$ \\
    \hline
     Low-Resource & +13.62\% & +17.74\%  & +0.16\% \\
     High-Resource & +29.38\% & +0.81\% & +0.41\% \\
     
     
     
    \hline
    
    \end{tabular}
  \end{adjustbox}
\caption{Refinement Analysis (LAS relative error reduction) of the UDify+RNGTr model for different refinement steps on the development sets of UD Treebanks.}
\label{table:refine}
\end{table}

To illustrate the refinement procedure for different dataset types, we split UD Treebanks based on their training set size into "Low-Resource" and "High-Resource" datasets.\footnote{We consider languages that have training data more than 10k sentences as "High-Resource".}
Table~\ref{table:refine} shows the refinement metric (${\tt REL}^t$) after each refinement iteration of the UDify+RNGTr model on these sets of UD Treebanks.\footnote{For these results we apply MST decoding after every iteration, to allow proper evaluation of the intermediate graphs.}  Every refinement step achieves an increase in accuracy, on both low and high resource languages.  But the amount of improvement generally decreases for higher refinement iterations.  Interestingly, for languages with less training data, the model cannot learn to make all corrections in a single step but can learn to make the remaining corrections in a second step, resulting in approximately the same total percentage of errors corrected as for high resource languages.
In general, different numbers of iterations may be necessary for different datasets, allowing efficiency gains by not performing unnecessary refinement iterations.

\begin{table}
\centering
  \begin{adjustbox}{width=\linewidth}
    \begin{tabular}{|l|c|c|c|}
    \hline
    Dependency Type & $t=1$ & $t=2$ & $t=3$ \\
    \hline
{\tt goeswith } & +57.83\% & +0.00\%  & +2.61\% \\ [-0.5ex]
{\tt aux } & +66.04\% & +3.04\%  & +3.12\%  \\ [-0.5ex]
{\tt cop } & +48.17\% & +2.21\%  & +3.01\%  \\ [-0.5ex]
{\tt mark } & +44.97\% & +2.44\%  & +0.00\%  \\ [-0.5ex]
{\tt amod } & +45.58\% & +2.33\%  & +0.00\% \\ [-0.5ex] 
{\tt det } & +34.48\% & +0.00\%  & +2.63\%  \\ [-0.5ex]
{\tt acl } & +33.01\% & +0.89\%  & +0.00\%  \\ [-0.5ex]
{\tt xcomp } & +33.33\% & +0.80\%  & +0.00\%  \\ [-0.5ex]
{\tt nummod } & +28.50\% & +0.00\%  & +1.43\% \\ [-0.5ex]
{\tt advcl } & +29.53\% & +1.26\%  & +0.25\%  \\ [-0.5ex]
{\tt dep } & +22.48\% & +2.02\%  & +0.37\%  \\ 
    \hline
    \end{tabular}
  \end{adjustbox}
\caption{Relative F-score error reduction of a selection of dependency types for each refinement step on the concatenation of UD Treebanks (with UDify as the initial parser).
\label{table:acc-deptype}}
\end{table}

\paragraph{Dependency Type Refinement:}
Table~\ref{table:acc-deptype} shows the relative improvement of different dependency types for the UDify+RNGTr model at each refinement step, ranked and selected by the total relative error reduction.
A huge amount of improvements is achieved for all these dependency types at the first iteration step, and then we have a considerable further improvement for many of the remaining refinement steps.
The later refinement steps are particularly useful for idiosyncratic dependencies which require a more global view of the sentence, such as auxiliary (aux) and copula (cop).
A similar pattern of improvements is found when SynTr is used as the initial parser, reported in Appendix~\ref{apx:syntr-refine}.

\begin{table}
\centering
  \begin{adjustbox}{width=\linewidth}
    \begin{tabular}{|l|c|c|c|}
    \hline
    Tree Type & $t=1$ & $t=2$ & $t=3$ \\
    \hline
    Non-Projective & +22.43\% & +3.92\% & +0.77\% \\
    Projective & +29.6\% & +1.13\% & +0.0\% \\
    \hline
    \end{tabular}
  \end{adjustbox}
\caption{Relative F-score error reduction of projective and non-projective trees on the concatenation of UD Treebanks (with UDify as the initial parser).
}
\label{table:pr-non-proj}
\end{table}

\paragraph{Refinement by Projectivity:}
Table~\ref{table:pr-non-proj} shows the relative improvement of each refinement step for projective and non-projective trees.  Although the total gain is slightly higher for projective trees, non-projective trees require more iterations to achieve the best results.  Presumably, this is because non-projective trees have more complex non-local interactions between dependencies, which requires more refinement iterations to fix incorrect dependencies.  This seems to contradict the common belief that non-projective parsing is better done with factorised graph-based models, which do not model these interactions.

%% file: conclusion.tex
\section{Conclusion}

In this article, we propose a novel architecture for structured prediction, Recursive Non-autoregressive Graph-to-Graph Transformer (RNG Transformer), to iteratively refine arbitrary graphs.  Given an initial graph, RNG Transformer learns to predict a corrected graph over the same set of nodes.  Each iteration of refinement predicts the edges of the graph in a non-autoregressive fashion, but conditions these predictions on the entire graph from the previous iteration.  This graph conditioning and prediction are made with the Graph-to-Graph Transformer architecture \cite{mohammadshahi2019graphtograph}, which can capture complex patterns of interdependencies between graph edges and can exploit BERT~\cite{devlin2018bert} pre-training.

We evaluate the RNG Transformer architecture by applying it to the problematic structured prediction task of syntactic dependency parsing.
In the process, we also propose a graph-based dependency parser (SynTr), which is the same as one iteration of our RNG Transformer model but without graph inputs.
Evaluating on
13 languages of the Universal Dependencies Treebanks, the English and Chinese Penn Treebanks, and the German CoNLL 2009 shared task treebank,
our SynTr model already significantly outperforms previous state-of-the-art models on all these treebanks.  Even with this powerful initial parser, RNG Transformer refinement almost always improves accuracies, setting new state-of-the-art accuracies for all treebanks.  RNG Transformer consistently results in improvement regardless of the initial parser,
reaching around the same level of accuracy even when it is given an empty initial parse, demonstrating the power of this iterative refinement method.
Error analysis suggests that RNG Transformer refinement is particularly useful for complex interdependencies in the output structure. 

The RNG Transformer architecture is a very general and powerful method for structured prediction, which could easily be applied to other NLP tasks.  It would especially benefit tasks that require capturing complex structured interdependencies between graph edges, without losing the computational benefits of a non-autoregressive model.

%% file: ack.tex
\section*{Acknowledgement}

We are grateful to the Swiss NSF, grant CRSII5\_180320,
for funding this work.
We also thank Lesly Miculicich, other members of the Idiap NLU group, the anonymous reviewers, and Yue Zhang for helpful discussions and suggestions.

%% file: supplementary.tex
\setcounter{section}{0}
\onecolumn
\begin{appendices}

\section{Implementation Details}
\label{apx:hyper-param}

For better convergence, we use two different optimisers for pre-trained parameters and randomly initialised parameters. We apply bucketed batching, grouping sentences by their lengths into the same batch to speed up the training. Early stopping (based on LAS) is used during training. We use "bert-multilingual-cased" for UD Treebanks.\footnote{\url{https://github.com/google-research/bert}. For Chinese and Japanese, we use pre-trained "bert-base-chinese" and "bert-base-japanese" models~\cite{Wolf2019HuggingFacesTS} respectively.} For English Penn Treebank, we use "bert-base-uncased", and for Chinese Penn Treebank, we use "bert-base-chinese". We apply pre-trained weights of "bert-base-german-cased"~\cite{Wolf2019HuggingFacesTS} for German CoNLL shared task 2009.
Here is the list of hyper-parameters for RNG Transformer model:

\begin{table}[hbt!]
    \begin{minipage}{.5\linewidth}
      \centering
        \begin{tabular}{c|c}
      Component & Specification \\
      \hline
      \textbf{Optimiser} & BertAdam \\
      Base Learning rate & 2e-3 \\
      BERT Learning rate & 1e-5 \\
      Adam Betas($b_1$,$b_2$) & (0.9,0.999) \\
      Adam Epsilon & 1e-5 \\
      Weight Decay & 0.01 \\
      Max-Grad-Norm & 1 \\
      Warm-up & 0.01 \\
      \hline
      \textbf{Self-Attention} \\
      No. Layers & 12 \\
      No. Heads & 12 \\
      Embedding size & 768 \\
      Max Position Embedding & 512 \\
      \hline
        \end{tabular}
        \end{minipage}
    \begin{minipage}{.5\linewidth}
      \centering
    \begin{tabular}{c|c}
      Component & Specification \\
      \hline
      \textbf{Feed-Forward layers (arc)} \\
      No. Layers & 2 \\
      Hidden size & 500 \\
      Drop-out & 0.33 \\
      Negative Slope & 0.1 \\
      \textbf{Feed-Forward layers (rel)} \\
      No. Layers & 2 \\
      Hidden size & 100 \\
      Drop-out & 0.33 \\
      Negative Slope & 0.1 \\
      \hline
      Epoch & 200 \\
      Patience & 100 \\
      \hline
    \end{tabular}
    \end{minipage} 
    \caption{\label{suptab:wsjhyper} Hyper-parameters for training on all Treebanks. We stop training, if there is no improvement in the current epoch, and the number of the current epoch is bigger than the summation of last checkpoint and "Patience".}
\end{table}

\section{Number of Parameters and Run Time Details:}
\label{apx:runtime}

We provide average run times and the number of parameters of each model on English Penn Treebanks, and English UD Treebank. All experiments are computed with a graphics processing unit (GPU), specifically the NVIDIA V100 model. We leave the issue of improving run times to future work.
\FloatBarrier
\begin{table}[hbt!]
	\centering
	\begin{adjustbox}{width=0.9\textwidth}
	\begin{tabular}{|@{~}l@{~}|@{~}c@{~}|@{~}c@{~}|@{~}c@{~}|}
		\hline
		Model & No. parameters & Training time (HH:MM:SS) & Evaluation time (seconds) \\
        \hline
		Biaffine~{\small\cite{dozat2016deep}} & 13.5M & 4:39:18 & 3.1 \\
		RNGTr & 206.3M & 24:10:40 & 20.6 \\
		SynTr & 206.2M & 6:56:40 & 7.5 \\
		\hline
	    \end{tabular}
	\end{adjustbox}
	\caption{Run time details of our models on English Penn Treebank.
        } 
\end{table}
\FloatBarrier
\begin{table}[hbt!]
	\centering
	\begin{adjustbox}{width=0.8\textwidth}
	\begin{tabular}{|l|c|c|}
		\hline
		Model & Training time (HH:MM:SS) & Evaluation time (seconds) \\
        \hline
		UDify~\cite{Kondratyuk_2019} & 2:22:47 & 4.0 \\
		RNGTr & 8:14:26 & 13.6 \\
		SynTr & 1:29:43 & 3.7 \\
		\hline
	    \end{tabular}
	\end{adjustbox}
	\caption{Run time details of our models on English UD Treebank.
        } 
\end{table}

\pagebreak
\section{Unlabelled Attachment Scores for UD Treebanks}

\label{apx:uas-ud}
\FloatBarrier
\begin{table}[hbt!]
  \centering
  \begin{tabular}{|c|c|cl|cl|c|}
    \hline
    \multirow{2}{*}{Language} &
    Multi & Multi & Multi+Mono & Mono & Mono & Mono \\
    & UDPipe & UDify & UDify+RNGTr & SynTr & SynTr+RNGTr & Empty+RNGTr \\
    \hline
    Arabic & 87.54 & 87.72 & \textbf{89.73}(+16.37\%) & \textbf{89.89} & \textbf{89.94}(+0.49\%) & \textbf{89.68} \\
    
    Basque & 86.11 & 84.94 & 90.49(+36.85\%) & 90.46 & \textbf{90.90}(+4.61\%) & \textbf{90.69} \\
    
    Chinese & 84.64 & 87.93 & 91.04(+25.76\%) & 91.38 & \textbf{92.47}(+12.64\%) & 91.81\\
    
    English & 89.63 & 90.96 & 92.81(+20.46\%) & \textbf{92.92} & \textbf{93.08}(+2.26\%) & 92.77 \\
    
    Finnish & 89.88 & 86.42 & \textbf{93.49}(+52.06\%) & \textbf{93.52} & \textbf{93.55}(+0.47\%) & \textbf{93.36} \\
    
    Hebrew & 89.70 & 91.63 & 93.03(+16.73\%) & \textbf{93.36} & \textbf{93.36}(0.0\%) & 92.80 \\
    
    Hindi & 94.85 & 95.13 & 96.44(+26.9\%) & 96.33 & \textbf{96.56}(+6.27\%) & 96.37 \\
    
    Italian & 93.49 & 95.54 & 95.72(+4.04\%) & \textbf{96.03} & \textbf{96.10}(+1.76\%) & \textbf{95.98}\\
    
    Japanese & 95.06  & 94.37 & \textbf{96.25}(+33.40\%) & \textbf{96.43} & \textbf{96.54}(+3.08\%) & \textbf{96.37} \\
    
    Korean & 87.70 & 82.74 & \textbf{91.32}(+49.71\%) & \textbf{91.35} & \textbf{91.49}(+1.62\%) & \textbf{91.28} \\
    
    Russian & 93.80 & 94.83 & \textbf{95.54}(+13.73\%) & \textbf{95.53} & \textbf{95.47}(-1.34\%) & \textbf{95.38} \\
    
    Swedish & 89.63 & 91.91 & 93.72(+22.37\%) & 93.79 & \textbf{94.14}(+5,64\%) & \textbf{94.14} \\
    
    Turkish & 74.19 & 74.56 & 77.74(+12.5\%) & \textbf{77.98} & \textbf{78.50}(+2.37\%) & 77.49 \\
    \hline
    Average & 88.94 & 89.13 & 92.10 & 92.23 & 92.46 & 92.16\\
    \hline
  \end{tabular}
  \vspace{-1ex}
    \caption{\label{uas-ud-table} Unlabelled attachment scores on UD Treebanks for monolingual (SynTr) and multilingual (UDPipe~\cite{straka-2018-udpipe} and UDify~\cite{Kondratyuk_2019}) baselines, and the refined models (+RNGTr), pre-trained with BERT \cite{devlin2018bert}.  Bold scores are not significantly different from the best score in that row (with $\alpha=0.01$).}
\end{table}

\section{SynTr Refinement Analysis}
\label{apx:syntr-refine}

\FloatBarrier
\begin{table}[hbt!]
\centering
  \begin{adjustbox}{width=0.45\linewidth}
    \begin{tabular}{|l|c|c|c|}
    \hline
    Dependency Type & $t=1$ & $t=2$ & $t=3$ \\
    \hline
{\tt clf } & +17.60\% & +0.00\%  & +0.00\% \\ 
{\tt discourse } & +9.70\% & +0.00\%  & +0.00\% \\ 
{\tt aux } & +3.57\% & +3.71\%  & +0.00\% \\ 
{\tt case } & +2.78\% & +2.86\%  & +0.00\% \\ 
{\tt root } & +2.27\% & +2.33\%  & +0.00\% \\ 
{\tt nummod } & +2.68\% & +1.38\%  & +0.00\% \\ 
{\tt acl } & +3.74\% & +0.29\%  & +0.00\%  \\ 
{\tt orphan } & +1.98\% & +1.24\%  & +0.00\% \\ 
{\tt dep } & +1.99\% & +0.80\%  & +0.00\% \\
{\tt cop } & +1.55\% & +0.78\%  & +0.00\% \\ 
{\tt advcl } & +1.98\% & +0.25\%  & +0.00\%  \\ 
{\tt nsubj } & +1.07\% & +0.54\%  & +0.00\%  \\ 
    \hline
    \end{tabular}
  \end{adjustbox}
  \vspace{-1ex}
\caption{Relative F-score error reduction, when SynTr is the initial parser, of different dependency types for each refinement step on the concatenation of UD Treebanks, ranked and selected by the total relative error reduction.
}
\end{table}

\begin{table*}[!ht]
\centering
  \begin{tabular}{|c|c|c|c|}
    \hline
    Dataset Type & $t=1$ & $t=2$ & $t=3$ \\
    \hline
     Low-Resource & 2.46\% & 0.09\%  & 0.08\% \\
     High-Resource & 0.81\% & 0.80\% & 0.32\% \\
    \hline
    \multicolumn{4}{c}{(a)}
    \end{tabular}
  ~~~
    \begin{tabular}{|l|c|c|c|}
    \hline
    Tree type & $t=1$ & $t=2$ & $t=3$ \\
    \hline
    Non-Projective  & 5\% & 1.63\% & 0.13\% \\
    Projective  & 0.6\% & 0.61\% & 0.13\% \\
    
    \hline
    \multicolumn{4}{c}{(b)}
    \end{tabular}
  \vspace{-1ex}
  \caption{
  Refinement analysis of the SynTr+RNGTr model for different refinement steps.
  (a) Relative LAS error reduction on the low-resource and high-resource subsets of UD Treebanks.
  (b) Relative F-score error reduction of projective and non-projective trees on the concatenation of UD Treebanks.
}
\end{table*}

\end{appendices}